\documentclass{article}
\usepackage{spconf,amsmath,graphicx}
\usepackage{times}
\usepackage{epsfig}
\usepackage{amsmath}
\usepackage{amssymb}
\usepackage{booktabs}
\usepackage{float} 
\usepackage{subfig}
\usepackage{algorithm}
\usepackage[noend]{algpseudocode}
\usepackage{hyperref}

\makeatletter
\newcommand{\paragraphh}{\@startsection{paragraph}{4}{0ex}%
   {-3.25ex plus -1ex minus -0.2ex}%
   {1.5ex plus 0.2ex}%
   {\normalfont\normalsize\bfseries}}
\makeatother

\title{Learning to Generate 3D Representations of Building Roofs Using Single-View Aerial Imagery}
\name{Maxim Khomiakov, Alejandro Valverde Mahou, Alba Reinders Sánchez, Jes Frellsen$^*$\thanks{$^*$ equal contribution}, Michael Riis Andersen$^*$}
\address{Technical University of Denmark}

\begin{document}
\maketitle

\begin{abstract}
We present a novel pipeline for learning the conditional distribution of a building roof mesh given pixels from an aerial image, under the assumption that roof geometry follows a set of regular patterns. Unlike alternative methods that require multiple images of the same object, our approach enables estimating 3D roof meshes using only a single image for predictions. The approach employs the PolyGen, a deep generative transformer architecture for 3D meshes. We apply this model in a new domain and investigate the sensitivity of the image resolution. We propose a novel metric to evaluate the performance of the inferred meshes, and our results show that the model is robust even at lower resolutions, while qualitatively producing realistic representations for out-of-distribution samples.
\end{abstract}

\section{Introduction}
\label{sec:intro}

Remote assessment or planning is necessary in various contexts, such as photovoltaic planning processes that require determining the optimal placement of solar panels on a roof surface. The combination of slope, azimuth and available area, are important inputs for solar panel project pricing, as well as solar potential production estimates \cite{kumar1997modelling,redweik2013solar}. Attaining these measures remotely, however, is not trivial. The motivation for this work is then to use a remote sensing medium in a setting that resembles an actual production environment, in particular we want to learn a conditional mapping from an aerial image to a 3D mesh representation of the roof, with just one image per building example. Prior studies have relied on using parameterized primitive shapes \cite{xiong2015flexible,alidoost20192d}, multi-view reconstruction or performing reconstruction with the addition of a direct depth modality, such as LiDAR.

Our approach offers a unique solution to the problem by using a single aerial image to learn the mapping, without requiring a direct depth modality or multiple images of the same object. We hypothesize that this is possible due to the inherent regularity of roof types, as well as the structural symmetry underlying common roof types, as previously demonstrated in studies such as \cite{Roof-class1, roof-class2}.

Estimating roof meshes from a single RGB image is attractive because such images are commonly available, unlike point cloud data. However, this task is typically considered ill-posed when performed using only a single image from a top-down perspective. Nevertheless, we propose that the inherent symmetries between the linear surfaces of building roofs enable us to learn a likely generative mapping. In high-resolution aerial imagery, roof surfaces are clearly visible, providing an opportunity to learn the conditional probability of a roof mesh for a given image. Our contributions can be summarized as follows: (1) We demonstrate the ability to learn a generative mapping from single-view images to a 3D roof mesh. (2) We present experimental results of our method under different image input conditions. (3) We introduce a novel measure of angular dissimilarity between triangular meshes and (4) demonstrate qualitative results from our model in an out of distribution setting.

\begin{figure}
    \centering
    \includegraphics[width=\linewidth]{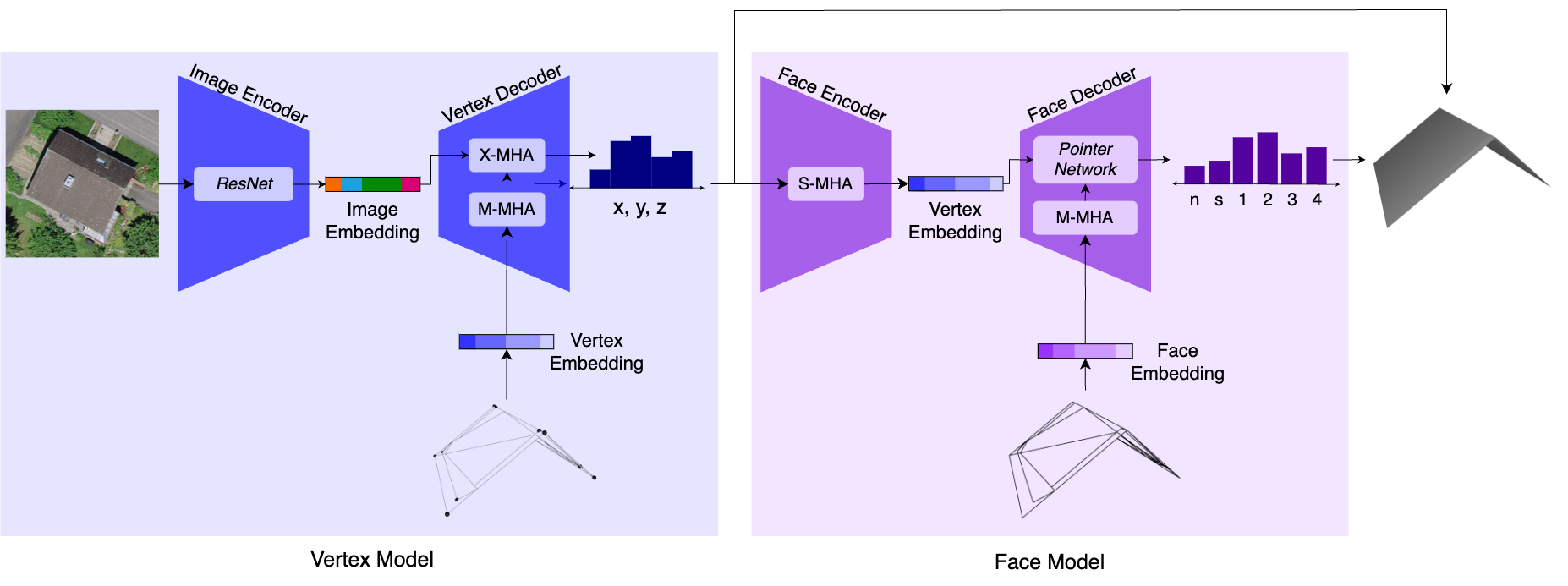}
    \caption{The model consists of two Encoder-Decoders, one for the input image and vertex embeddings, and another for the inferred vertices and face embeddings.}
    \label{fig:pipeline}
\end{figure}

\section{Related Work}
\label{sec:relatedwork}

\noindent \textbf{Photogrammetry} \;
Photogrammetry commonly involves using imagery to create a depth modality from image input. For 3D building reconstruction, several studies propose using LiDAR, aerial images, or a combination of both, as seen in \cite{lidarroof, facade}. Hu \textit{et al.}~\cite{roofreconstruction} present an approach that uses a multi-label energy minimization solution and a roof connection graph to semantically decompose and reconstruct LiDAR point clouds into complex roof structures.

Our approach is unique in that we learn a generative mapping from a single image during training, enabling us to learn structures that are not directly visible from the input image, unlike other approaches that require multiple images or modalities. 

\vspace{.5em}
\noindent \textbf{RNN \& CNNs} \; Prior research has proposed several models for generating 3D objects from RGB images, mostly based on CNNs and RNNs, as seen in \cite{multiviewdl}. For instance, the 3D-R2N2 model \cite{3D-R2N2} reconstructs 3D objects based on one or more images from randomly chosen viewpoints, using a 3D-ConvLSTM architecture that models voxels. Another model, Pix2Vox \cite{pix2vox}, employs an encoder-decoder architecture with 2D convolutional image encoders, 3D convolutional decoders, and a refiner auto-encoder to generate the final 3D volume. These models employ varying views of the object during training and optimize for voxel representations that are scale-dependent. In contrast, our approach predicts meshes that are directly applicable in downstream tasks.

Pixel2Mesh++ \cite{Pixel2Mesh++} generates 3D shapes of triangular meshes from a single RGB image by performing progressive deformations on an ellipsoid, utilizing features extracted from the image by a 2D CNN. This approach has been extended to produce a mesh from multiple images. Although it allows inference from a single image, it is trained by using multiple viewpoints of the same object class, which is not necessary in our approach.

\vspace{.5em}
\noindent \textbf{Transformers} \; Transformers have demonstrated remarkable performance improvements in speech and translation tasks, surpassing the performance of RNNs, as seen in \cite{TransvsRNN1, TransvsRNN2}. Several studies have applied Transformers to 3D reconstruction, including 3D-RETR \cite{multiviewreconstruction3D-RETR} and VolT \cite{multiview3d}, which both utilize multi-view inputs and generate 3D voxel outputs. Notably, 3D-RETR can reconstruct 3D objects from a single view. However, similar to other methods we compare against, these approaches rely on training with multiple views of the same object class, while our approach only requires a single view without class labels. 

\vspace{.5em}
\noindent \textbf{Other methods} \; Neural radiance fields are a popular method for 3D reconstruction that often rely on multiple images. Recent studies, however, demonstrate the ability to learn novel viewpoints without the need for multiple images by using self-supervised learning techniques with pseudolabels and geometric regularizations \cite{Xu_2022_SinNeRF}. Another study \cite{tucker2020single} demonstrates view synthesis from a single image input at inference time, but relies on videos with a moving camera during training. In contrast, our approach differs in several ways. First, we impose fewer constraints. Second, we do not assume access to unseen views nor multiple images of the same object. Third, we do not focus on texture or depth, but instead consider the problem as a sequence-to-sequence task, treating vertices and faces as next-token predictions in 3D space.

\section{Method}
\label{sec:method}

The goal is to reconstruct a mesh $\mathcal{M} = {(\mathcal{V},\mathcal{F})}$ for a given image $\mathcal{I}$. This task can be formulated as learning the conitional distribution  $p(\mathcal{M}|\mathcal{I})$ using the decomposition
\begin{align}
p(\mathcal{M}|\mathcal{I}) & = p(\mathcal{V}, \mathcal{F}|\mathcal{I}) \label{eqn:distribution_1} \\
&= p(\mathcal{F}|\mathcal{V,I})p(\mathcal{V}|\mathcal{I})
\end{align}
where $p(\mathcal{V}|\mathcal{I})$ is the distribution of the vertices $\mathcal{V}$ given the image, and $p(\mathcal{F}|\mathcal{V,I})$ is the distribution of the set of faces $\mathcal{F}$ conditioned on the vertices and the image.

The PolyGen \cite{polygen} architecture models these quantities by parameterizing the distributions using a deep neural network consisting of CNNs and Transformer modules. Transformers model a sequence of tokens using multi-head attention (MHA) to learn the dependencies between the tokens. Since MHA is permutation invariant, we need to use some form of encoding prior to feeding our sequence to the model. For this work, we rely on four types of learned embeddings: a dimensional embedding $E_\textrm{dim}$ encoding the $x$, $y$, or $z$ dimension in the sequence, a positional embedding $E_\textrm{pos}$ which learns the order in which tokens appear in the sequence, and a token embedding (face or vertex value) $E_\textrm{token}$. To produce a parsing of the image $E_\textrm{image}$, we use a pre-trained ResNet encoder and add a fixed value to each pixel of the feature representation from the final convolutional layer. We then perform a linear transformation to match the dimension of the hidden state of the vertex transformer decoder.

\vspace{.5em}
\noindent \textbf{Vertex Model} \;
The objective of the vertex model is to express a distribution over sequences of vertices. The vertices are ordered from lowest to highest by focusing on the $z$-coordinate, as such we concatenate tuples of $(z_i, y_i, x_i)_i$ coordinates to obtain a flattened sequence. The joint distribution over the vertex sequence $\mathcal{V}^\textrm{seq}$ is modeled autoregressively as the product of a series of conditional vertex distributions
\begin{align}
\label{eqn:vertex}
p(\mathcal{V}^\textrm{seq}|\mathcal{I}, \theta_{v}) = \prod_{n=1}^{N_v} p(v_n|v_{<n}, \mathcal{I} ,\theta_{v}),
\end{align}
where $\theta_{v}$ is the parameters of the model. The distribution (Eqn. \ref{eqn:vertex}) models the tokens as a function of prior predicted tokens at each time step defined over an 8-bit quantized vertex space with a stopping token $s$ to know when the sequence is complete. The encoder produces image embeddings that consists of a set of ResNet \cite{ResNet} blocks producing a feature map being embedded with an image coordinate and linearly projected to the hidden dimensions of the encoder.

\vspace{.5em}
\noindent \textbf{Face Model} \; The face model represents a distribution over a sequence of mesh faces, which is conditioned on the mesh vertices predicted by the vertex model. Similar to the vertices, the faces are ordered with the lowest index first, and the faces $(f_1^{(i)},f_2^{(i)}, \ldots, f_{N_i}^{(i)})_i$ are concatenated to form a flattened sequence.

As done in the vertex model, the distribution is outputted over the values of $f$ at each set, and is trained by maximizing the log-likelihood of $\theta_{f}$ over the training set. This is a categorical distribution, and it is defined by the predicted vertices and two special tokens: an end-of-face token $n$ and an end-of-sequence token $s$.
\begin{align}
    p(\mathcal{F}^\textrm{seq}|\mathcal{V}; \theta_{f}) = \prod_{n=1}^{N_F} p(f_n|f_{<n}, \mathcal{V}, \theta_{f}).
\end{align}
The target distribution $p(\mathcal{F}^\textrm{seq}|\mathcal{V}; \theta_{f})$ is defined over the indices of an input set of vertices, which varies depending on the batch, and therefore a PointerNet architecture \cite{pointer} is used.%

The decoder is a Transformer decoder that works on sequences of face token embeddings. It is conditioned on the input vertices via dynamic face embeddings and through cross-attention into the sequence of vertex embeddings. The embeddings used are similar to the vertex model, $E_\textrm{dim}$, $E_\textrm{pos}$ and $E_\textrm{token}$.

\vspace{.5em}
\noindent \textbf{Angular Dissimilarity} \; Algorithm \ref{alg:angular_dissimilarity} calculates the angular dissimilarity between two surfaces, $\mathbf{A}$ and $\mathbf{B}$. First, the normal vectors of each surface are computed, denoted by sets $\mathcal{A}$ and $\mathcal{B}$, respectively. For each surface normal in set $\mathcal{A}$, the minimum angle between it and set $\mathcal{B}$ is computed and added to the running angle measure, $\angle A$. This process is repeated for set $\mathcal{B}$, and the final score is normalized by the number of faces in each set.

\begin{algorithm}[tbp]
\caption{Angular Dissimilarity}\label{alg:angular_dissimilarity}
\begin{algorithmic}[1]
\small{
\Procedure{Compute Angular Dissimilarity}{$\mathcal{A},\mathcal{B}$}
\State $\angle A,\angle B \leftarrow 0$
\For{$\forall a_i \in \mathcal{A}$}
        \State $\angle A \leftarrow \angle A + \min\left[\angle(a_i,\mathcal{B})\right]$
      \EndFor{end for} 
\For{$\forall b_j \in \mathcal{B}$}
        \State $\angle B \leftarrow \angle B + \min\left[\angle(b_j,\mathcal{A})\right]$
      \EndFor{end for} \\
    $\angle AB \leftarrow \frac{\angle A}{2|\mathcal{A}|} + \frac{\angle B}{2|\mathcal{B}|}$ \\
    return $\angle AB$
\EndProcedure
}
\end{algorithmic}
\end{algorithm}

\section{Experiments}
\label{sec:experiments}

We evaluate our approach through experiments using two datasets, one containing watertight roof surfaces and another containing non-watertight surfaces. Since there are no applicable benchmarks to compare our results to, we use appropriate baselines and metrics to measure performance, including the PoLiS distance \cite{polisDistance}, intersection over union (IoU), and the angular dissimilarity metric (Algorithm \ref{alg:angular_dissimilarity}). The PoLiS distance measures the degree of overlap between the vertices of the estimated and ground truth surfaces, the IoU-score measures object localization accuracy in pixel space, and the angular dissimilarity metric provides a measure of coplanar cohesion for the estimated surfaces.  %

\noindent \textbf{Data} \; The datasets used in this study are sourced from swisstopo, the Federal Office of Topography of Switzerland \cite{swisstopo}, and comprise pairs of images and 3D meshes. The images are obtained from aerial orthoimagery, using the bounding box of the corresponding mesh and a 3-meter buffer on each side to ensure complete building visibility. The meshes were acquired using photogrammetric methods for 3D reconstruction with stereo aerial imagery. We constrain our examples to a maximum vertex sequence length of 100 and no more than 400 faces. The resulting dataset contains 159,412 examples for the watertight roof set and 697,025 examples for the non-watertight set. The datasets were split into training, validation, and test sets, with a distribution of 70\%, 15\%, and 15\%, respectively.

\vspace{.5em}
\noindent \textbf{Training} \;
We train our model using the Adam optimizer with a learning rate of $2 \cdot 10^{-5}$, $\beta_{1}=0.9$, and $\beta_{2}=0.99$. We use a dropout rate of 0.4 in the image encoder, 0.3 in the vertex decoder and face encoder, and 0.2 in the face decoder. We apply early stopping after 25 epochs for the \textit{PolyGen} model and 15 epochs for the \textit{PolyGen (w)} model. The vertex decoder comprises five layers, each with four attention heads, and embeddings of dimension 512 with fully connected layers of 1024. The face encoder consists of three layers with four attention heads and fully connected layers of size 512. The face decoder comprises four layers with four attention heads and fully connected layers of size 512.

\subsection{Results}

\begin{figure}[tbp]%
    \centering
    \subfloat[\centering PoLiS]{{\includegraphics[width=.48\linewidth]{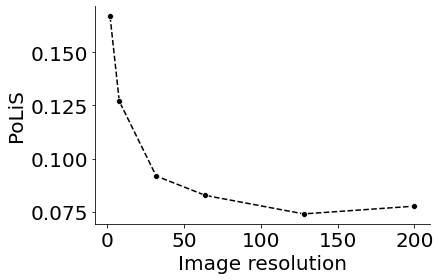} }}%
    \subfloat[\centering Angular Dissimiarily]{{\includegraphics[width=.48\linewidth]{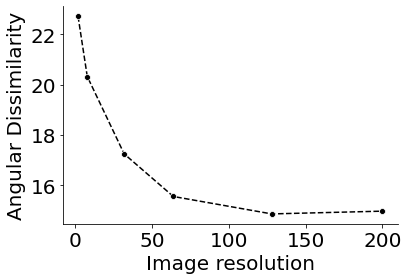} }}%
    \caption{Average PoLiS and Angular Dissimilarity conditioned on test-set images for models trained with varying resolutions. The largest standard deviation of the mean was in the order of $10^{-8}$ and $10^{-2}$ for PoLiS and Angular Dissimiarily consequtively.}%
    \label{fig:comparison_ablation}%
\end{figure}

\begin{figure}[t!]
    \centering
    \includegraphics[width=.8\linewidth]{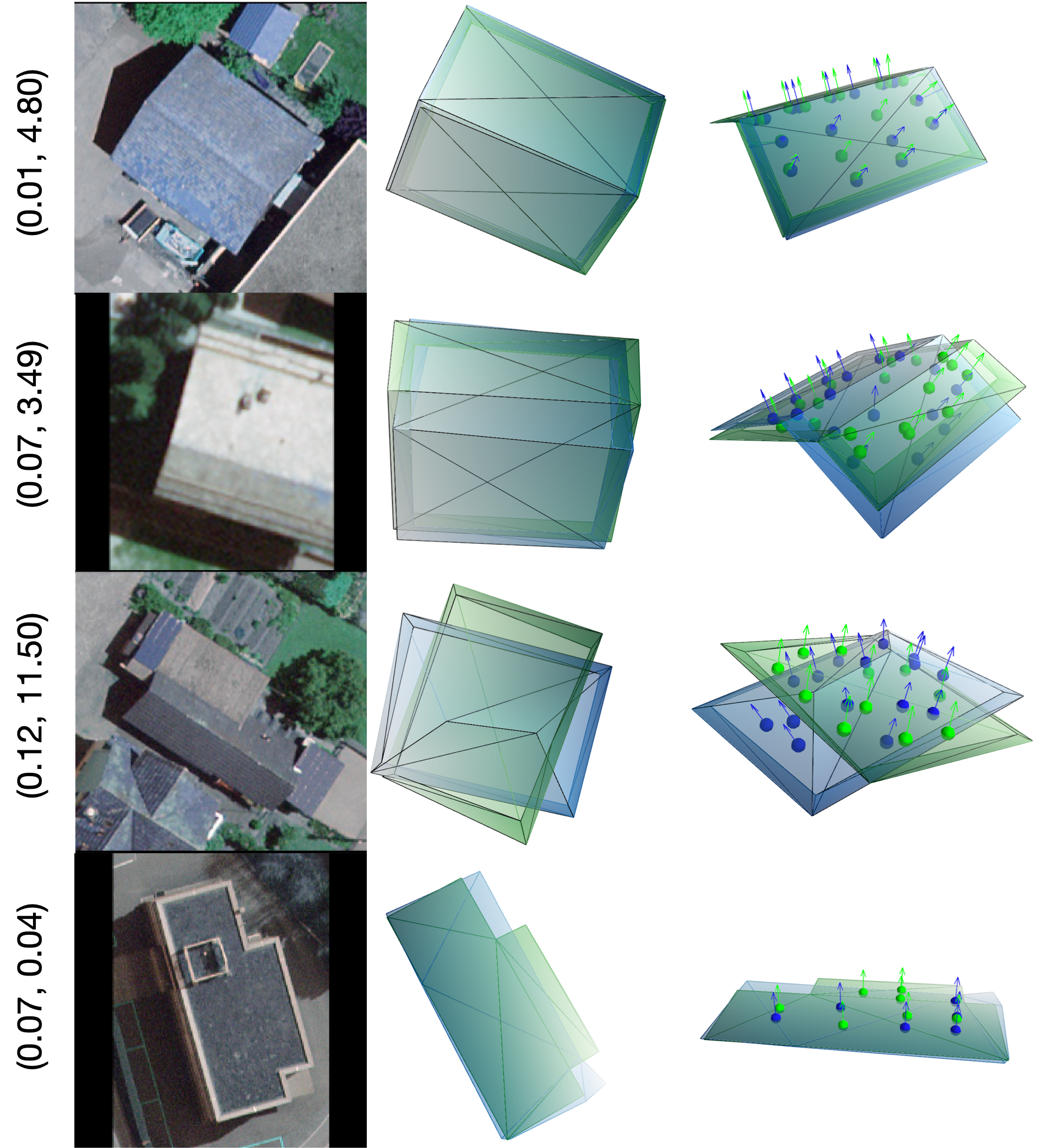}
    \caption{The predicted examples from the test set. The PoLiS distance and angular dissimilarity values for each example are shown in parentheses, while the input image used for conditioning is displayed in the first column. The second and third columns illustrate the PoLiS distance and angular dissimilarity visualizations, respectively, with the ground truth values shown in green and the predicted values in blue.}
    \label{fig:results_inference}
\end{figure}

We present our results in Table \ref{tab:results1}. Nucleus Sampling \cite{https://doi.org/10.48550/arxiv.1904.09751} was performed with a probability of $p=0.95$ for all predicted examples. We computed metrics on all examples from both datasets (i.e., watertight and non-watertight meshes), except for the PolyGen* scenario, where we constrained the inference computation to a sample of 28,000 test images to save time. A set of generated samples can be seen in Figure \ref{fig:results_inference}. We observe a clear connection between a low PoLiS distance and angular dissimilarity, and the quality of the generated mesh. Similarly, in Figure \ref{fig:comparison_ablation}, we note how the model is surprisingly robust to changes in image resolution.

\begin{figure}[t!]
    \centering
    \includegraphics[width=.8\linewidth]{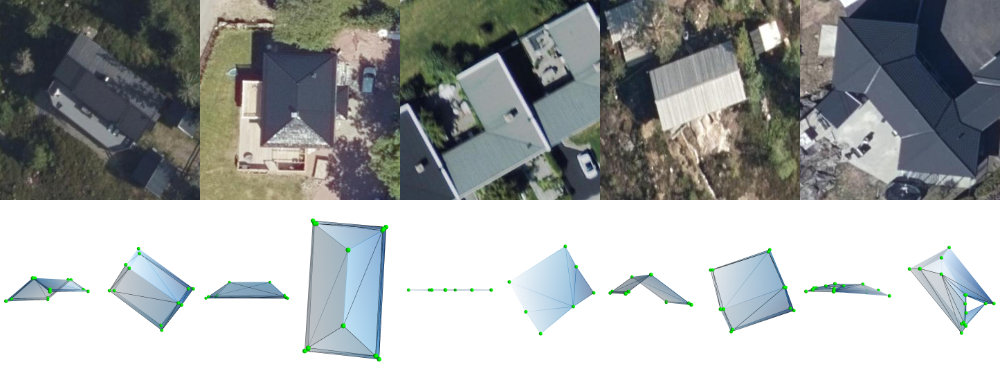}
    \caption{Predicted examples using images from Norkart in Norway (out of distribution inference). Each example has two figures of the predicted roof mesh: The left image is shown from an azimuthal angle of $0_{xy}$° and $90_{z}$° zenith elevation, while the right is shown with an azimuth and elevation of 0°.}
    \label{fig:inthewild_inference}
\end{figure}

\vspace{.5em}
\noindent \textbf{Baseline calculations} \;
We evaluate the performance having two baselines. A Random-case, where for every \textit{test}-mesh we choose a random training-mesh as the ground truth, and a ResNet baseline, where we use a pretrained ResNet-101 to attain a feature vector for each image. We use these vectors to train a KNN-model, using the features corresponding to the training set using the nearest predicted test-sample as ground truth for the baseline computation.

\vspace{.5em}
\noindent \textbf{Inference on images in the Wild} \;
Figure \ref{fig:inthewild_inference} demonstrates our model's ability to perform inference on images captured from a geographically distant location and acquired by a different camera. The Norkart data shares characteristics with the SwissTopo dataset, having a similar ground sampling distance and the presence of a single building centered in the image. Although the sample from Norkart in Norway is small and purely qualitative, our model shows to produce generally plausible predictions of building roofs for these images in the wild, the model fails however, for images with more than one building, or if the building has complicated roof topology.

\begin{table}[tbp]
\centering
{%
\caption{Results benchmarked against test dataset. (w) Indicates measures on the watertight dataset. PolyGen* indicates inference on a subset of the test dataset.}
\label{tab:results1}
\begin{tabular}{@{}llll@{}} \toprule
\emph{Method} & $\downarrow$ PoLiS  & $\downarrow$ $\angle AB$ & $\uparrow$ IoU   \\ \midrule
Random       &  0.231  & 28.8 & 0.19  \\
ResNet            & 0.197   & 25.1 & 0.27  \\ 
PolyGen*  &     \textbf{0.089}   & \textbf{24.1} & \textbf{0.47}  \\ 
\midrule
Random (w)      & 0.257 & 32.4 & 0.19  \\
ResNet (w)    &    0.208         & 28.9 & 0.26  \\
PolyGen (w)         &  \textbf{0.078}           & \textbf{15.3} & \textbf{0.78}  \\\bottomrule
\end{tabular}%
}
\end{table}

\section{Conclusions}
\label{sec:conclusions}

We have presented a novel approach to generating 3D meshes of building roofs from their corresponding aerial images. Our hypothesis is based on the regular symmetrical relations found on building roofs, suggesting that it may not be necessary to use a direct depth modality for 3D roof reconstruction, given access to a large dataset. Our results demonstrate that the method can be applied to a more complex domain than originally intended, and that the model is able to generate a 3D roof mesh conditioned on a single aerial image from a fixed viewpoint. We believe that the potential of this approach has yet to be fully explored, and it may prove useful in various industrial purposes, particularly in photovoltaic deployment and planning.

\section{Acknowledgements}
We extend our thanks to Otovo AS for funding this research.

\bibliographystyle{IEEEbib}
\bibliography{strings,refs}

\begin{thebibliography}{10}

\bibitem{kumar1997modelling}
Lalit Kumar, Andrew~K Skidmore, and Edmund Knowles,
\newblock ``Modelling topographic variation in solar radiation in a gis
  environment,''
\newblock {\em International journal of geographical information science}, vol.
  11, no. 5, pp. 475--497, 1997.

\bibitem{redweik2013solar}
Paula Redweik, Cristina Catita, and Miguel Brito,
\newblock ``Solar energy potential on roofs and facades in an urban
  landscape,''
\newblock {\em Solar energy}, vol. 97, pp. 332--341, 2013.

\bibitem{xiong2015flexible}
Biao Xiong, M~Jancosek, S~Oude Elberink, and G~Vosselman,
\newblock ``Flexible building primitives for 3d building modeling,''
\newblock {\em ISPRS Journal of Photogrammetry and Remote Sensing}, vol. 101,
  pp. 275--290, 2015.

\bibitem{alidoost20192d}
Fatemeh Alidoost, Hossein Arefi, and Federico Tombari,
\newblock ``2d image-to-3d model: Knowledge-based 3d building reconstruction
  (3dbr) using single aerial images and convolutional neural networks (cnns),''
\newblock {\em Remote Sensing}, vol. 11, no. 19, pp. 2219, 2019.

\bibitem{Roof-class1}
M.~Buyukdemircioglu, R.~Can, and S.~Kocaman,
\newblock ``Deep learning based roof type classification using very high
  resolution aerial imagery,''
\newblock {\em The International Archives of the Photogrammetry, Remote Sensing
  and Spatial Information Sciences}, vol. XLIII-B3-2021, pp. 55--60, 2021.

\bibitem{roof-class2}
Jeremy Castagno and Ella Atkins,
\newblock ``Roof shape classification from lidar and satellite image data
  fusion using supervised learning,''
\newblock {\em Sensors}, vol. 18, no. 11, 2018.

\bibitem{lidarroof}
Jaewook Jung and Gunho Sohn,
\newblock ``A line-based progressive refinement of 3d rooftop models using
  airborne lidar data with single view imagery,''
\newblock {\em ISPRS Journal of Photogrammetry and Remote Sensing}, vol. 149,
  pp. 157--175, 03 2019.

\bibitem{facade}
Yunsheng Zhang, Chi Zhang, Siyang Chen, and Xueye Chen,
\newblock ``Automatic reconstruction of building façade model from
  photogrammetric mesh model,''
\newblock {\em Remote Sensing}, vol. 13, no. 19, 2021.

\bibitem{roofreconstruction}
Pingbo Hu, Yiming Miao, and Miaole Hou,
\newblock ``Reconstruction of complex roof semantic structures from 3d point
  clouds using local convexity and consistency,''
\newblock {\em Remote Sensing}, vol. 13, no. 10, 2021.

\bibitem{multiviewdl}
Dawen Yu, Shunping Ji, Jin Liu, and Shiqing Wei,
\newblock ``Automatic 3d building reconstruction from multi-view aerial images
  with deep learning,''
\newblock {\em ISPRS Journal of Photogrammetry and Remote Sensing}, vol. 171,
  pp. 155--170, 2021.

\bibitem{3D-R2N2}
Christopher~B. Choy, Danfei Xu, JunYoung Gwak, Kevin Chen, and Silvio Savarese,
\newblock ``3d-r2n2: A unified approach for single and multi-view 3d object
  reconstruction,''
\newblock {\em Lecture Notes in Computer Science}, p. 628–644, 2016.

\bibitem{pix2vox}
Haozhe Xie, Hongxun Yao, Xiaoshuai Sun, Shangchen Zhou, and Shengping Zhang,
\newblock ``Pix2vox: Context-aware 3d reconstruction from single and multi-view
  images,''
\newblock {\em 2019 IEEE/CVF International Conference on Computer Vision
  (ICCV)}, Oct 2019.

\bibitem{Pixel2Mesh++}
Chao Wen, Yinda Zhang, Zhuwen Li, and Yanwei Fu,
\newblock ``Pixel2mesh++: Multi-view 3d mesh generation via deformation,''
  2019.

\bibitem{TransvsRNN1}
Shigeki Karita, Nanxin Chen, Tomoki Hayashi, Takaaki Hori, Hirofumi Inaguma,
  Ziyan Jiang, Masao Someki, Nelson Enrique~Yalta Soplin, Ryuichi Yamamoto,
  Xiaofei Wang, Shinji Watanabe, Takenori Yoshimura, and Wangyou Zhang,
\newblock ``A comparative study on transformer vs {RNN} in speech
  applications,''
\newblock in {\em 2019 {IEEE} Automatic Speech Recognition and Understanding
  Workshop ({ASRU})}. dec 2019, {IEEE}.

\bibitem{TransvsRNN2}
Surafel~M. Lakew, Mauro Cettolo, and Marcello Federico,
\newblock ``A comparison of transformer and recurrent neural networks on
  multilingual neural machine translation,'' 2018.

\bibitem{multiviewreconstruction3D-RETR}
Zai Shi, Zhao Meng, Yiran Xing, Yunpu Ma, and Roger Wattenhofer,
\newblock ``3d-retr: End-to-end single and multi-view 3d reconstruction with
  transformers,''
\newblock 2021.

\bibitem{multiview3d}
Dan Wang, Xinrui Cui, Xun Chen, Zhengxia Zou, Tianyang Shi, Septimiu Salcudean,
  Z.~Jane Wang, and Rabab Ward,
\newblock ``Multi-view 3d reconstruction with transformer,'' 2021.

\bibitem{Xu_2022_SinNeRF}
Dejia Xu, Yifan Jiang, Peihao Wang, Zhiwen Fan, Humphrey Shi, and Zhangyang
  Wang,
\newblock ``Sinnerf: Training neural radiance fields on complex scenes from a
  single image,''
\newblock 2022.

\bibitem{tucker2020single}
Richard Tucker and Noah Snavely,
\newblock ``Single-view view synthesis with multiplane images,''
\newblock in {\em Proceedings of the IEEE/CVF Conference on Computer Vision and
  Pattern Recognition}, 2020, pp. 551--560.

\bibitem{polygen}
Charlie Nash, Yaroslav Ganin, S.~M.~Ali Eslami, and Peter~W. Battaglia,
\newblock ``Polygen: An autoregressive generative model of 3d meshes,'' 2020.

\bibitem{ResNet}
Kaiming He, Xiangyu Zhang, Shaoqing Ren, and Jian Sun,
\newblock ``Deep residual learning for image recognition,'' 2015.

\bibitem{pointer}
Navdeep~Jaitly Oriol~Vinyals, Meire~Fortunato,
\newblock ``Pointer networks,'' 2017.

\bibitem{polisDistance}
Janja Avbelj, Rupert Müller, and Richard Bamler,
\newblock ``A metric for polygon comparison and building extraction
  evaluation,''
\newblock {\em IEEE Geoscience and Remote Sensing Letters}, vol. 12, no. 1, pp.
  170--174, 2015.

\bibitem{swisstopo}
Federal~Office of~Topography~swisstopo,
\newblock ``Federal office of topography swisstopo,''
  \url{https://www.swisstopo.admin.ch/en/home.html}.

\bibitem{https://doi.org/10.48550/arxiv.1904.09751}
Ari Holtzman, Jan Buys, Li~Du, Maxwell Forbes, and Yejin Choi,
\newblock ``The curious case of neural text degeneration,'' 2019.

\end{thebibliography}

\end{document}